\documentclass[conference]{IEEEtran}

\usepackage{graphicx}
\usepackage{graphics} 
\usepackage{epsfig} 
\usepackage{mathptmx} 
\usepackage{times} 
\usepackage{amsmath} 
\usepackage{amssymb}  
\usepackage{caption}
\usepackage{subcaption}
\usepackage{float}
\floatstyle{plaintop}
\restylefloat{table}
\usepackage[tableposition=top]{}
\restylefloat{table}
\usepackage{url}

\newcommand{\RNum}[1]{\uppercase\expandafter{\romannumeral #1\relax}}

\begin{document}


%
\title{Improved Microaneurysm Detection using Deep Neural Networks}

\author{\IEEEauthorblockN{Mrinal Haloi}
\IEEEauthorblockA{
Indian Institute of Technology, Guwahati\\
Email: h.mrinal@iitg.ernet.in/mrinal.haloi11@gmail.com}}


%


\maketitle

\begin{abstract}

In this work, we propose a novel microaneurysm (MA) detection for early diabetic retinopathy screening using color fundus images. Since MA usually the first lesions to appear as an indicator of diabetic retinopathy, accurate detection of MA is necessary for treatment. Each pixel of the image is classified as either MA or non-MA using a deep neural network with dropout training procedure using maxout activation function. No preprocessing step or manual feature extraction is required. Substantial improvements over standard MA detection method based on the pipeline of preprocessing, feature extraction, classification followed by post processing is achieved. The presented method is evaluated in publicly available Retinopathy Online Challenge (ROC) and Diaretdb1v2 database and achieved state-of-the-art accuracy.\\


\end{abstract}


\IEEEpeerreviewmaketitle

\section{Introduction}
In recent days diabetic retinopathy (DR) is one of the most common severe eye diseases causing blindness in developing and developed countries. According to WHO \cite{c14} DR is the primary pathology for 4.8\% of the 37 million blindness cases around the world. Since DR is a progressive disease, early stage detection and treatment can save the patient from losing sight. For analyzing progress in disease fundus image of patient need be checked regularly. Fast and reliable automatic computer aided diagnosis system will reduce the burden on specialists and will give better performance for DR mass screening.  In most of the DR screening system sensitivity and specificity is used as efficiency measurement.

\begin{figure}
  \centering
      \includegraphics[width=3.1in,height=2.8in]{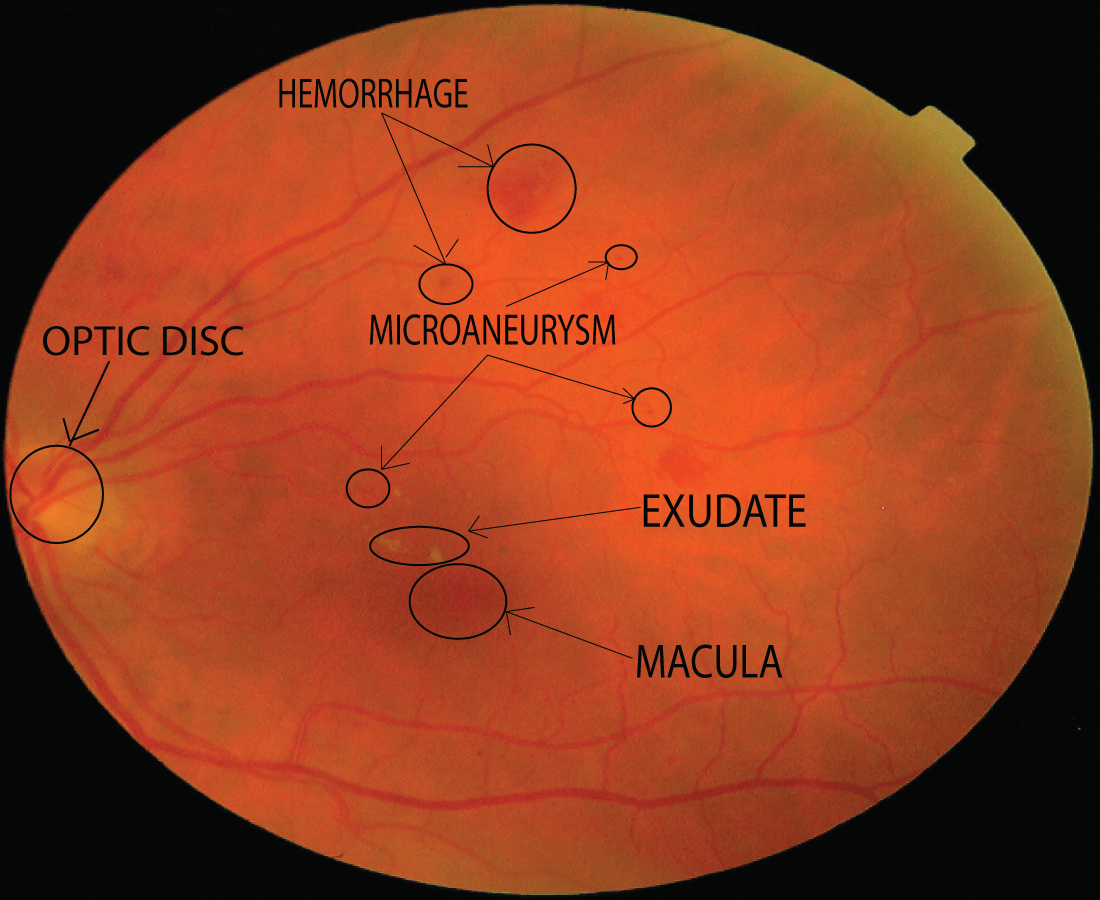}
\caption{Typical Pathological Retinal Image}
\label{fig:over}
\end{figure}

In general, MA appears as the first lesson for diabetic retinopathy. Reliable detection of MA has major importance for diabetics screening purpose. In color fundus images MA appears as small red dots with the very small radius less than that of the major optic vain. In reality, these are tiny swollen capillaries in the retina, can discharge blood leading to other pathological symptoms such as exudates, hemorrhages etc.  Various challenges such as vessels bifurcations and crossing, illumination and contrast changes, artifacts, degradation of the image due to imaging device setup etc. appear in automatic fundus image based DR screening system. A full proof DR screening system is capable of the detection of clinical features such as exudate, MA, hemorrhages, cotton wool spots and blood vessel damages. A recent state-of-the-art method for exudate and cotton wool spots detection was presented by Haloi et al. \cite{haloi}. The MA detection is a well-investigated research area for DR mass screening system. Our motivation of these works is to present a new method to detect MA under the different challenging situation and to achieve high sensitivity and specificity. Fig.~\ref{fig:over} depicts typical retinal image with pathological features such as microaneurysms, exudates etc and non-pathological features such as the optic disc, the macula etc. The complexity of MA detection can be observed from Fig.~\ref{fig:over}.

Most of the MA detection works presented till now have a common pipeline of three to four stages; first preprocessing the image, secondly manual feature extraction followed by classification and a postprocessing step. Also, the use of the high contrast green channel of the fundus  image is very common in MA detection research. In general, existing methods used a morphological method, filtering based and supervised classification using hand crafted features etc. Antal et al. \cite{c1} had developed ensemble-based microaneurysms detection system and claimed first prize in ROC online challenge and also achieved a good result on another dataset.

They have ensemble several preprocessing and candidate extraction method to develope their final model. But they didn't addressed the problems of degradation and illuminance changes. Also Quellec et al. \cite{c7} Proposed template matching-based method for MA detection in wavelet domain and developed optimally adapted wavelet family. Their method prone false detection and true rejection due to haemorrhages and big vessels respectively. Neijmer et al. \cite{c9} combined previously existed method for candidate extraction and used pixel-wise classification using manually designed features. In another works by pereira et al. \cite{c10} exploited multi-agent system for MA segmentation preceded by gaussian and kirsch filter based preprocessing. Final MA candidates evolves from multi-agent interaction with preprocessed image. Lazer et al. \cite{c8} microaneurysms was detected using rotating cross section profiles based method which depends on circularity and diameter of MAs. For each profiles peak was detected and features such as size, height and shape was calculated.

In this work, we propose a deep learning based pixel-wise MA classification method invariant to luminance, contrast changes and artifacts. No image based preprocessing or feature extraction stages is required. In addition to that this method performance independent on vessel structures, the optic disc and the fovea. Hence extraction or detection of these features are not required. To increase the accuracy of the method dropout training with maxout activation function is used. Training of this network is time consuming but testing phase is very fast and suitable for real-time applications. We have achieved state-of-the-art performance with a very low false positive rate on publicly available datasets.

\section{Methods}
\begin{figure}
  \centering
      \includegraphics[width=3.3in,height=1.8in]{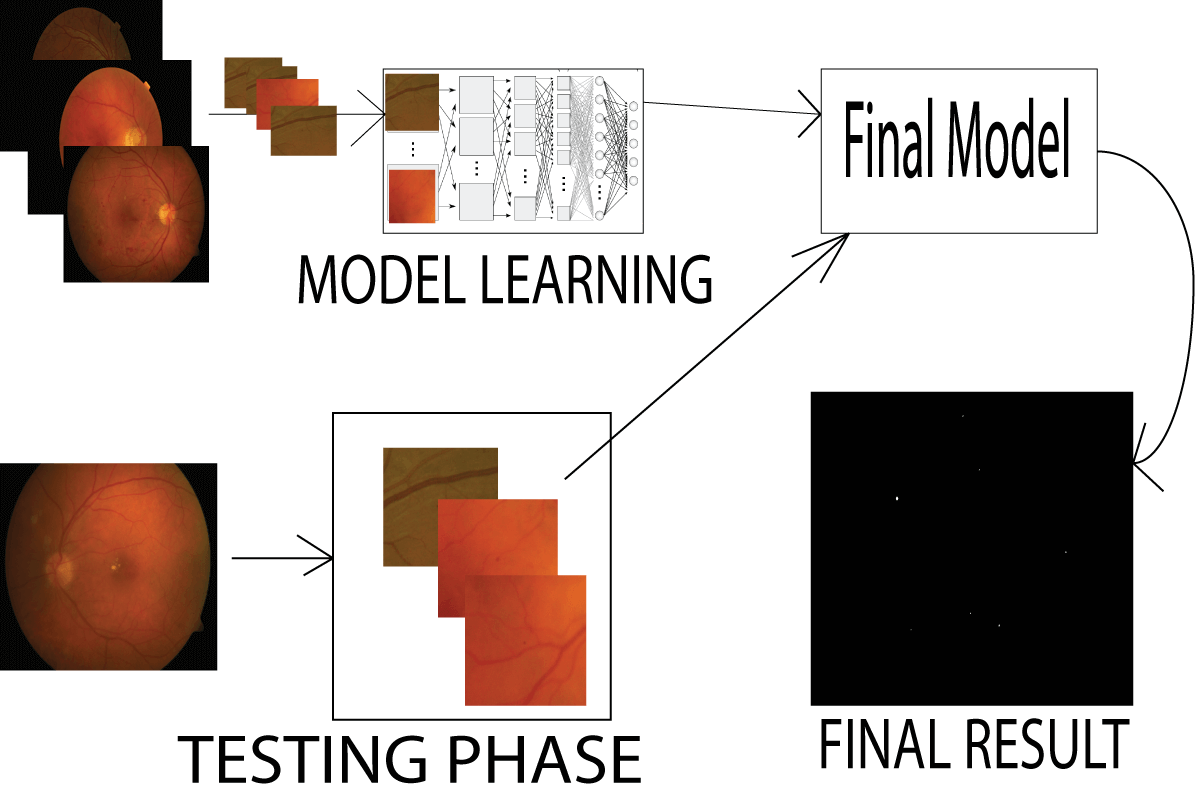}
\caption{Method Overview}
\label{fig:algo}
\end{figure}

Microaneurysms (MA) usually follow a gaussian-like intensity distribution and have isolated structures from neighbours. To detect this tiny structures a pixel based deep neural network (DNN) \cite{dnn} is developed. Pixel based classification is useful for this type of complex detection. Every pixel of the image is classified as MA or non-MA. For any given pixel, class label is predicted using three color channel RGB values in a square window centered on that pixel of size w. The window around the given pixel may contain other MA. An overview of the method has been depicted in Fig.~\ref{fig:algo}. 

\subsection{Data Manipulation}
Because of local maximum structures of MA; rest of pixels of the window centered on that pixel need to be processed efficiently to get high classification probability. For the account of this effect input data was modified by using two techniques for neighbouring data suppression, specifically foveation and nonuniform sampling. The concept of foveation originated from uneven size and organization of photo-receptive cells and ganglions in the human eye. Visual acuity is maximum in the middle of the retina termed as fovea and decreases towards the periphery of the retina. Foveation proved to be very effectively in nonlocal means denoising algorithms \cite{fov}. In foveation central section of the window is focused, while the peripheral pixels are defocused using linear space invariant gaussian blur. The standard deviation of gaussian blur kernels increases with distance from the central section. \\

It has been observed that increasing input window size in DNN improves performance significantly, but at the same computational time complexity also increases. Nonuniform sampling was used to selectively depreciate window pixels towards the periphery. 
Only central section of input window is sampled at full resolution, while sampling resolution decreases towards the periphery. Using this method large window can be trained with relatively fewer neurons.

\subsection{Network Architecture}
DNNs are hierarchical neural networks, inspired by the simple and complex cells in the human primary visual cortex. A DNN comprised of convolutional layer alternate wth max-pooling layer \cite{maxpool} followed by fully connected layers and a final classification layer. DNN very definite power of learning discriminative features from raw image patches make it efficient for computer vision tasks, in comparisons to traditional handcrafting features. The network used in this work contains five layers including the classification layer; the first three are comprised of convolutional layers each followed by max-pooling. The convolutional layers are followed by one fully connected hidden layers and the softmax classification layer is fully connected with two neurons for MA and non-MA. In this work, we have incorporated dropout \cite{dropout} training algorithm for three convolutional layers and one fully connected hidden layer. And maxout activation function is used for all layers in the network except the softmax layer.

\subsubsection{Convolutional Layer}
The convolutional layer \cite{cnn} is the core building block of a deep neural network parameterized by the input volume size $M_{i} \times M_{i} \times D_{i}$, the receptive filed or filter size $F$, the depth of conv layer $K$ and the stride or skipping factor $S$. If input border is zero padded with size of $P_{i}$ then number of neurons in the output volume $M_{o} \times M_{o} \times D_{o}$ in  is calculated as follows.

\begin{equation}
M_{o} = \frac{M_{i} - F + 2P}{S} +1; D_{o} = K
\end{equation}
Stride should be chosen such that $M_{o}$  is an integer.

\subsubsection{Max-Pooling Layer}
Max-pooling layer ensures fast convergence in comparison to traditional neural networks. In addition to that max -pooling provides translation invariance. The input image is partitioned into a set of non-overlapping rectangles and the maximum value of each subregion is chosen for output. If $W(k,l)$ are subregions then the output is obtained as follows. 

\begin{equation}
y_{k,l} = \max_{ij \in W(k,l)} x_{i,j}
\end{equation} 
Suppose the input volume of size $M_{i}\times M_{i} \times  D_{i}$ for max-pooling layer with spatial extent $F$ and skipping factor S; then output volume of size $M_{o}\times M_{o} \times D_{o}$ is calculated as follows.

\begin{equation}
M_{o} = \frac{M_{i} - F}{S} +1; D_{o} = D_{i}
\end{equation}
If value of $F > S$, then the process is called overlapping pooling, in general model with overlapping pooling is less prone to overfit. 

\subsubsection{Dropout and maxout}
Dropout \cite{dropout} is one of the most important improvements in machine learning, proved to be successful in many application. It has been observed that combining the output of many models improve accuracy significantly, nut in case of deep neural networks training many models more than computationally costly. Nitish et al. \cite{dropout} introduced dropout training for deep neural networks, means to reduce overfitting by randomly omitting the output of each hidden neuron with a probability of 0.5. Training is similar to the standard neural network using stochastic gradient descent. The only difference is that dropped out neurons don't take part in forward pass and backpropagation. Suppose a neural network model with $L$ hidden layers and $W$, $b$ are weights and biases matrix of the network. If $l \in {1,2,...,L}$ is hidden layer index; $z^{l}$ and $y^{l}$ denote vector of inputs and outputs respectively at layer l. The following equation described feed forward operation.
\begin{equation}
\begin{aligned}
z^{l+1} = W^{l+1}y^l + b^{l+1}\\
y^{l+1} = f(z^{l+1})
\end{aligned}
\end{equation}
Using dropout training, the feed forward equation becomes

\begin{equation}
\begin{aligned}
(r_{i})^{l} = Bernoulli(p) \\
\check{y^{l}} = r^{l} * y^{l}  \\
z^{l+1} = W^{l+1}\check{y^l} + b^{l+1}\\
y^{l+1} = f(z^{l+1})  \\
\end{aligned}
\end{equation}
where f any activation fucntion, in our case f is maxout activation function.\\

\begin{figure}
  \centering
      \includegraphics[width=3.1in,height=1.6in]{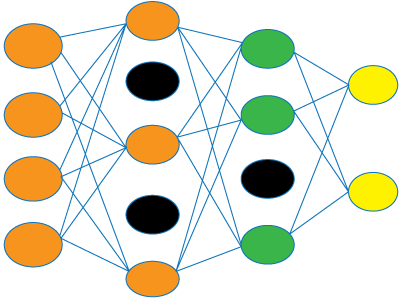}
\caption{Dropout training illustration}
\label{fig:dropout}
\end{figure}

A typical situation of dropout training has been explained in Fig.~\ref{fig:dropout}, the black circle denotes dropped out node from the network. Dropped out nodes do not participate in training and testing.

\begin{figure}
  \centering
      \includegraphics[width=3.1in,height=1.6in]{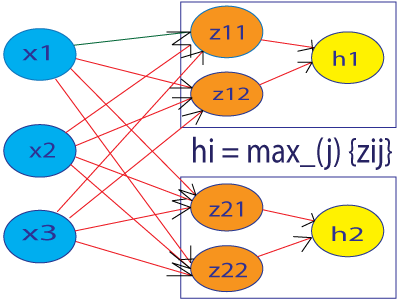}
\caption{Maxout activation function illustration}
\label{fig:maxout}
\end{figure}

The conventional way to represent a neuron's output $f$ as a function of its input x with $f(x) = (1 + e^{-x})^{-1})$ or $f(x) = tanh(x)$. Problems arise with this type of function in gradient descent training, as these functions saturate early with positive and negative $x$ values. Gradient descent stuck in this type of function, but lots better  improvement can be achieved slightly modifying the activation function as proposed by Goodfellow et al. \cite{maxout}, the maxout network. Maxout is a new kind of activation function for the deep neural network with dropout training procedure. In maxout algorithm, the input is divided into the activation function into $k$ unit groups and maximum response is recorded. Fig.~\ref{fig:maxout} depicts typical situation of maxout activation function.  
Given a input $x \in R^d$, a maxout hidden uint $h_{i}$ implements the following function
\begin{equation}
\begin{aligned}
z_{i,j} =  x^TW_{...ij} + b_{i,j}  \\
h_{i}(x) = \max_{j \in [1,k]}z_{i,j}
\end{aligned}
\end{equation}
where $W \in R^{d\times m \times k}$ and $b \in R^{m \times k}$.

\subsection{Learning nets}
 From pixel level expert annotated ground truth each pixel is considered either as MA or non-MA. The training set consists of windows centered on image pixels. If a window lies party outside of the image border, rest of the pixels are derived by horizontal reflections. Windows with a MA pixel at the center is considered as MA samples and that of with non-MA considered for non-MA samples for training. Moreover to reduce overfitting and to ensure rotational invariance the most common method is to enlarge the dataset using random rotations and using horizontal reflections for border pixels. 

The training procedure for dropout neural networks using maxout activation function resemblance with traditional neural networks except a few things. In the case of dropout network learning each neuron is dropped with a probability of $Bernouli(p)$, resulting a thinned network. In addition to that forward and backpropagation are done only on this thinned network. Convergence of stochastic gradient descent has got much better improvement in this network using maxout activation function. Also, one particular form of regularization specifically constraining the norm of the incoming weight vector at each hidden unit found to be especially useful for dropout training. This is termed as max-norm regularization inspired from the previous use in the context of collaborative filtering \cite{cola}.  
\begin{equation}
\begin{aligned}
\Delta w^{t} = m^{t}\Delta w^{t-1} - (1-m^{t})\epsilon^{t} <\nabla_{w}L> \\
w^{t} = w^{t-1} + \Delta w^{t} \\
\|w^{t}\| \leq c
\end{aligned}
\end{equation}

the weight norm constraint only for fully connected layers.
\begin{equation}
\begin{aligned}
\epsilon^{t} = \epsilon_{0}f^{t} \\
 m^{t}= 
\begin{cases}
    \frac{t}{T}m_{i} + (1 - \frac{t}{T})m_{f} &  t < T \\
    m_{f},              & t\geq T
\end{cases}
\end{aligned}
\end{equation}

where c is a fixed constant, t is the iteration index, $\epsilon$ is the learning rate,$m$ is the momentum variable.

\section{Experiments, Results and Discussions}
This method have been tested on publicly available ROC \cite{c11}, Messidor \cite{c12} and Diaretdb1v2 \cite{c13} dataset. Both of these well annotated with pixel-wise labelling, which facilitates the design of our pixel-wise classification model. ROC contain 50 training image of 768*576 pixels, Messidor consists of 1200 losslessly compressed images with 45 degrees field of view and Diaretdb1v2 includes 89 images of 1500*1152 pixels. The images of Messidor were captured using 8 bits per color plane at 1440*960, 2240*1488 or 2304*1536 pixels. Each image is provided with a grading score of R0 to R3. R0 and R1 correspond to no DR and mild DR respectively; where as R2 and R3 are sever DR and proliferate DR respectively. The grading based on a number of MAs and Haemorrhages with presence or absence of neovascularization. No grading scheme available for ROC and Diaretdb1v2 datasets. Not all the images of have MA as pathological features. Pixels with another label such as haemorrhages, blood vessels crossings (between two different vessels) and bifurcations (one vessel originated from another one) and end point of disconnected vessels are considered as non-MA samples. For each pixel, input network receives six different windows using data augmentation. For each pixel, three windows are obtained using vertical and horizontal mirroring. And then, each window was modified using foveation and nonuniform sampling producing two final windows, this setting emphasizes the central pixel and efficient use of bigger window size. Images were taken at different conditions with different cameras with their native resolution and compression settings. The retinal specialist annotations were obtained from a combination of three ophthalmologists with retinal fellowship training. 
All experiments are conducted on a Ubuntu machine with 12GB RAM, intel i7 3.10GHz processor, and NVIDIA GTX 590 graphics card with 1024 CUDA cores. We use Pylearn2 \cite{pylearn} machine learning library built on the top of Theano. Pylearn2 come with an efficient implementation of dropout training with maxout activation function. Total 90000 MA and 1.5 million non-MA windows were used to train the network. While constructing the non-MA windows it has been emphasised to include an extensive number of possible false positives. And a small number of trivial non-MA windows were included. This setting helps the network to learn proper distinctive features.

For accuracy analysis for exudates detection we will compute true positive (TP) a number of exudates pixels correctly detected, false positive (FP) a number of non-exudate pixels which are detected wrongly as exudate pixels, false negative (FN) number of exudate pixels that were not detected and true negative (TN) a number of no exudates pixels which were correctly identified as non-exudate pixels. For better representation of accuracy sensitivity and specificity at pixel level was used as our measurement. Thus the global sensitivity SE and the global specificity SP and accuracy AC for each image are defined as follows.

\begin{equation}
\begin{aligned}
SE = \frac{TP}{TP + FN}\\
PRED = \frac{TP}{TP + FP}\\
SP = \frac{TN}{TN + FP} \\
AC = \frac{TP + TN}{TP + TN+FP + FN}
\end{aligned}
\end{equation}

The best network architecture is depicted in Table. \RNum{1} with three convolutional layers each followed by a max-pooling layer and one fully connected layer. And a softmax layer on the top of the network with two neurons for MA and non-MA probability values. The probability of dropping a neuron on each layer is shown in $Bernouli(p)$ column. Each convolutional layer has size $5\times5$ with a stride of 2 pixels for the first layer and 1 pixel for next two layers. Overlapping pooling is used in each of the max-pooling layers with a stride of 2 pixels and pooling regions size $3\times 3$.

\begin{table}[H]
  \centering
  \begin{tabular}{*{20}{c}}
\hline
Layer & Type & Maps \& Neurons & Size & Stride & Bern(p)\\
\hline
0 & input & $3 \times 129 \times 129$ & ... & ...& 0.1 \\
\hline
1 & Conv & $64 \times 63 \times 63$ & $ 5 \times 5$ & 2 &  0.2 \\
\hline
2 & MP & $64 \times 31 \times 31$ & $ 3 \times 3$ & 2 & .. \\
\hline
3 & Conv & $64 \times 27 \times 27$ & $ 5 \times 5$ & 1 &  0.2 \\
\hline
4 & MP & $64 \times 13 \times 13$ & $ 3 \times 3$ & 2 & .. \\
\hline
5 & Conv & $64 \times 9 \times 9$ & $ 5 \times 5$ & 1 &  0.5 \\
\hline
6 & MP & $64 \times 4 \times 4$ & $ 3 \times 3$ & 2 & .. \\
\hline
7 & FC & 290 & $1 \times 1$ &... & 0.5\\
\hline
8 & FC & 2 & $1 \times 1$ &... & ..\\
\hline
\end{tabular} 
  \caption{Network Architecture}
\end{table}

To detect MA in an unseen image, we first apply a mask to get all pixels of interest removing the usual black region appears during fundus photography. Also, a color threshold was defined to left out trivial non-MA pixels to reduce computation time. The window of size $129 \times 129$ centered at each image pixel is extracted, for pixels nearby boundaries windows were extracted using horizontal mirroring. Each window has R, G, B color channel. The detector will assign a probability value of being MA and non-MA to each pixel in the image. Finally, a probability map of being MA is generated for the testing image.

To remove possible false detection each connected region of probability map is processed using the concept of convexity and area of the region. Let's consider M is the set of all MA pixels then  
\begin{itemize}

\item{N: Number of connected region in probaility Map.}

\item{For each region  update  $M = M \cup P_{R}| Area_{P_{R}} \leq 21 \cap Convexity_{P_{R}} \geq 0.8$ }  
\item{$Area_{P_{R}}$ is the area of the region $P_{R}$ and $Convexity_{P_{R}}$ is the convexity}

\end{itemize}

This will ensure that no vessels crossing, bifurcations and haemorrhages are included in MA detection.

\begin{figure}
  \centering
      \includegraphics[width=2.5in,height=2.5in]{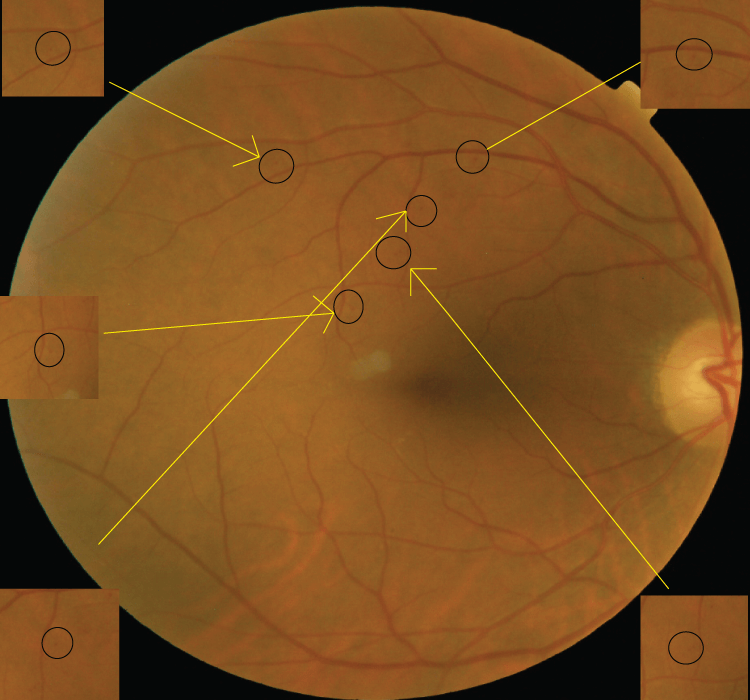}
\caption{Detected MA pixels at the center of the windows}
\label{fig:output}
\end{figure}
For a typical image pixel based detection output is shown in Fig.~\ref{fig:output}. We have observed that this method can reliably detect MA candidates.  

A comparison of this method with existing DR screening system is shown in Table \RNum{2}. Even though this comparison is not done on the same ground since dataset and the proportion of images having DR symptoms are different. But Sensitivity (Sens), Specificity (Spec) and area under the curve (AUC) value can be accepted for mutual comparison. Our method performs significantly better than the existing methods. 

\begin{table}[H]
  \centering
  \begin{tabular}{*{20}{c}}
\hline
Method & DR(\%) & Sens & Spec & AUC  \\
\hline
Proposed Method & 46 &  97\%   &   95\%    &  0.988       \\
\hline
Antal et al. \cite{c1} & 46 &  90\% & 91\%  & 0.989 \\
\hline
Agurto et al. \cite{c2} & 76.26 & N\/A & N\/A & 0.89	\\
\hline
Abramoff et al. \cite{c3} & 4.8 & 84\% & 64\% & 0.84 \\
\hline
Jelinek et al. \cite{c4} & 30 & 85\% & 90\% & N\/A    \\
\hline
\end{tabular} 
  \caption{Comparison of automatic DR screening systems.}
\end{table}

\begin{figure}
  \centering
      \includegraphics[width=3.3in,height=1.8in]{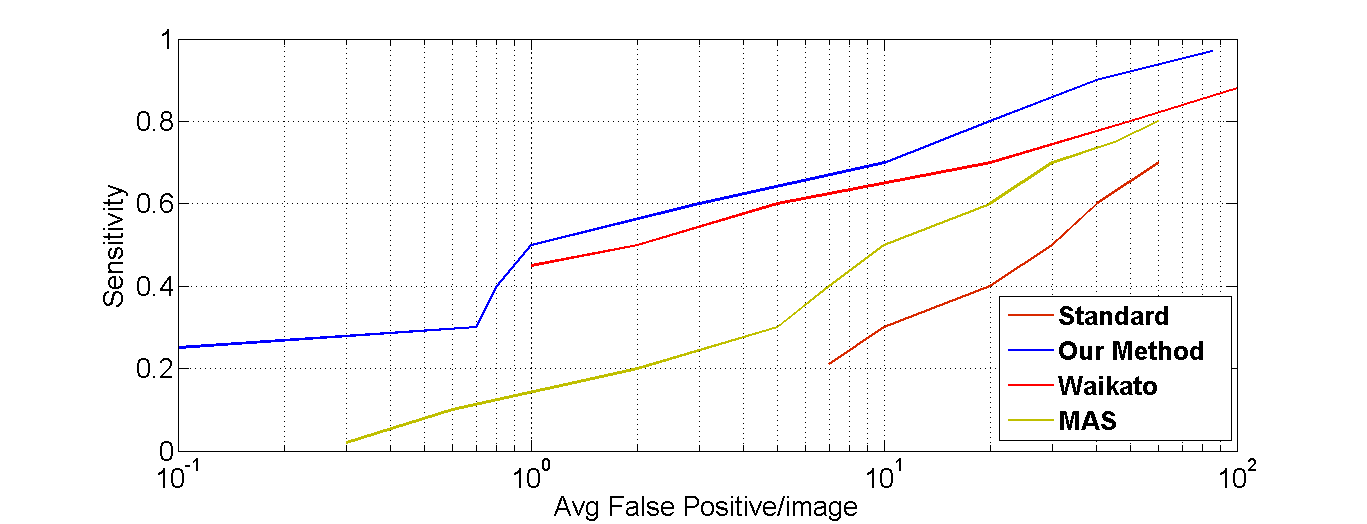}
\caption{Comparisons of Sensitivity vs Average Number False positive pixels per image}
\label{fig:sevsfp1}
\end{figure}
\begin{figure}
  \centering
      \includegraphics[width=3.3in,height=1.8in]{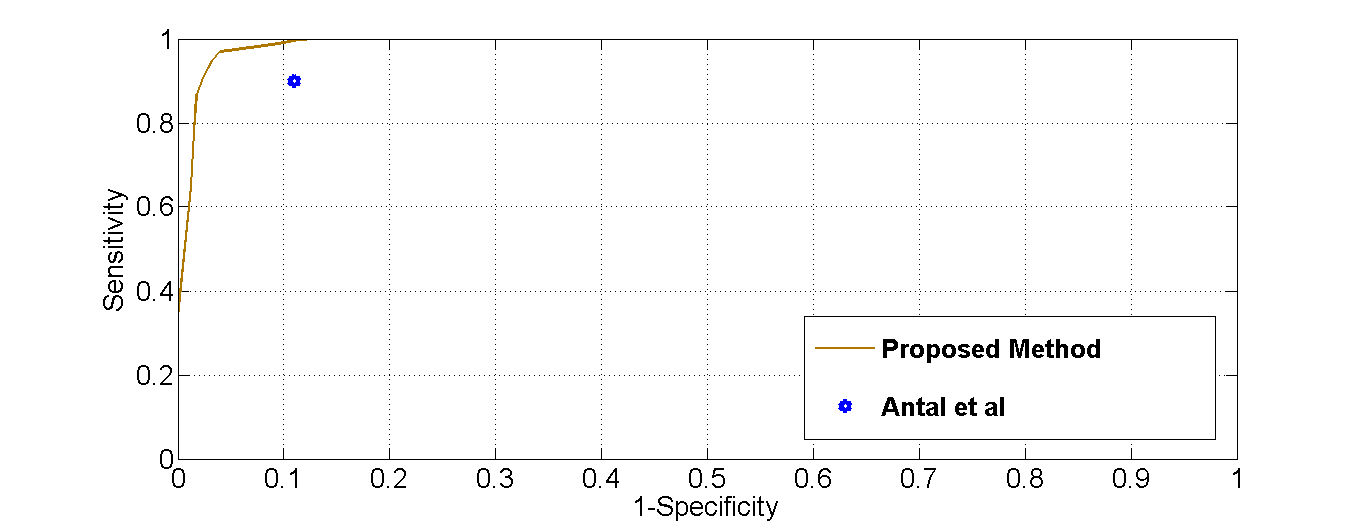}
\caption{Comparisons of Sensitivity vs 1-Specificity}
\label{fig:sevssp1}
\end{figure}

This method achieved lower false positive rate than other existing systems. A comparison of sensitivity vs an average number of false positive pixels per image is shown in Fig.~\ref{fig:sevsfp1}. Variations of sensitivity with 1-specificity are shown in Fig.~\ref{fig:sevssp1}. Also for comparison purpose result of one existing method also plotted in this figure. 

Table \RNum{3} shows a comparison of this method on Messidor dataset with recent state-of-the-art system for the scenario R0 vs R1. It can be observed that our method achieve a accuracy of 95\% with sensitivity and specificity of 97\% and 94\% respectively.

\begin{table}[H]
  \centering
  \begin{tabular}{*{20}{c}}
\hline
Method & Sensitivity & Specificity & Acc & AUC  \\
\hline
Proposed Method &   97\%   &   95\%    &   95.4\%    &  0.982       \\
\hline
Antal et al. \cite{c1} & 94\% & 90\% & 90\% & 0.942 \\
\hline

\end{tabular} 
  \caption{Comparisons of result on the Messidor Dataset for the scenario R0 vs R1}
\end{table}
Also for the scenario No DR vs DR Table \RNum{4} shows result comparisons with existing state-of-the-art method on the same dataset.

\begin{table}[H]
  \centering
  \begin{tabular}{*{20}{c}}
\hline
Method & Sensitivity & Specificity & Acc & AUC  \\
\hline
Proposed Method &   97\%   &   96\%    &   96\%    &  0.988       \\
\hline
Antal et al. \cite{c1} & 90\% & 91\% & 90\% & 0.989 \\
\hline

\end{tabular} 
  \caption{Comparisons of result on the Messidor Dataset for the scenario No DR/DR}
\end{table}

A extensive evaluation was also carried out on ROC dataset. Due to lack of testing data label we have used a part training data (not used in the training of this method) to evaluate accuracy of our method. Table \RNum{5} depicts comparison of AUC values with other methods on the same dataset. 

\begin{table}[H]
  \centering
  \begin{tabular}{*{20}{c}}
\hline
Method  & AUC  \\
\hline
Proposed Method  &  0.98       \\
\hline
Human Expert & 0.96 \\
\hline
OK Medical \cite{c5}  & 0.89 \\
\hline
Fujita Lab \cite{c6} & 0.88 \\
\hline
LaTIM \cite{c7} & 0.87 \\
\hline

\end{tabular} 
  \caption{Comparisons of result on the ROC Dataset (Our result only on subset of train Data, Since test data label not available)}
\end{table}

\section{Conclusion}
In this work, we have presented a deep learning based computer-aided system for microaneurysm detection. The deep network consists of 5 layers including softmax output layer and dropout training with maxout activation function is used to improve accuracy. In comparison to another existing method, this system does not require additional blood vessels extraction step, preprocessing and feature design. This method has been tested in publicly available datasets and  achieved state-of-the-art performance for MA candidates extraction with low false positive rate, hence useful for diabetic mass screening purpose.


\section*{Acknowledgment}

\end{document}